%% file: main.tex
\documentclass[10pt,twocolumn,letterpaper]{article}

\usepackage{cvpr}              

\newtheorem{theorem}{Theorem}[section]   

\newtheorem{definition}[theorem]{Definition}

\usepackage{booktabs}
\usepackage{multirow}
\usepackage{colortbl}
\usepackage{verbatim}
\usepackage{xcolor}
\usepackage{arydshln}

\input{preamble}
\definecolor{cvprblue}{rgb}{0.21,0.49,0.74}
\usepackage[pagebackref,breaklinks,colorlinks,allcolors=cvprblue]{hyperref}

\def\confName{CVPR}
\def\confYear{2026}

\title{CAMD: Coverage-Aware Multimodal Decoding for Efficient Reasoning of Multimodal Large Language Models}

\author{\textbf{Huijie Guo}\textsuperscript{\rm 1,\rm 2}\thanks{These authors contributed equally.}, 
    \textbf{Jingyao Wang}\textsuperscript{\rm 1,\rm 2}\footnotemark[1], 
    \textbf{Lingyu Si}\textsuperscript{\rm 1,\rm 2}, 
    \textbf{Jiahuan Zhou}\textsuperscript{\rm 3}, \\
    \textbf{Changwen Zheng}\textsuperscript{\rm 1,\rm 2}, 
    \textbf{Wenwen Qiang}\textsuperscript{\rm 1,\rm 2}\thanks{Corresponding author.}\and
    \textsuperscript{\rm 1}Institute of Software Chinese Academy of Sciences,
    \textsuperscript{\rm 2}University of Chinese Academy of Sciences,\\
    \textsuperscript{\rm 3}Wangxuan Institute of Computer Technology, Peking University
}

\begin{document}
\maketitle
\input{sec/0_abstract}
\input{sec/1_intro}
\input{sec/2_rework}

\input{sec/3_analysis}

\input{sec/4_methods}


\input{sec/5_exp}

\input{sec/6_conclusion}



{
    \small
    \bibliographystyle{ieeenat_fullname}
    \bibliography{main}
}


\end{document}

%% file: sec/0_abstract.tex
\begin{abstract}
Recent advances in Multimodal Large Language Models (MLLMs) have shown impressive reasoning capabilities across vision–language tasks, yet still face the challenge of compute–difficulty mismatch. Through empirical analyses, we identify that existing decoding methods may waste compute on easy cases while underserving hard ones, affecting both model effectiveness and efficiency. To address this issue, we first develop a theoretical framework that links sampling coverage, instance difficulty, and residual risk. Our analysis reveals that multimodal reasoning exhibits a heavy-tailed difficulty distribution; a small subset of hard or ambiguous samples dominates the residual failure probability. Based on this insight, we propose Coverage-Aware Multimodal Decoding (CAMD), an adaptive inference mechanism that dynamically allocates computation according to estimated uncertainty. CAMD integrates evidence-weighted scoring, posterior coverage estimation, and sequential Bayesian updating to balance efficiency and reliability under a limited token budget. Experiments on various benchmark datasets and baselines demonstrate the effectiveness and advantages of our approach.
\end{abstract}

%% file: sec/1_intro.tex
\section{Introduction}
\label{sec:intro}

Multimodal large language models (MLLMs)~\cite{zhu2023minigpt, zhang2023internlm, chen2023shikra, bai2023qwen, Instructblip, liu2024visual, liu2024improved, dong2024internlm} are increasingly used for vision tasks and visual question answering because they combine powerful content understanding~\cite{lai2024lisa} with generation capabilities~\cite{geng2024instructdiffusion}.
Compared with text-only LLMs, MLLMs must jointly solve semantic understanding and cross-modal alignment to produce coherent scene interpretations; this coordination across perception, language, and reasoning improves applicability but also raises computational cost and makes inference difficulty highly instance-dependent.

\begin{figure}[t]
  \centering
   \includegraphics[width=0.92\linewidth]{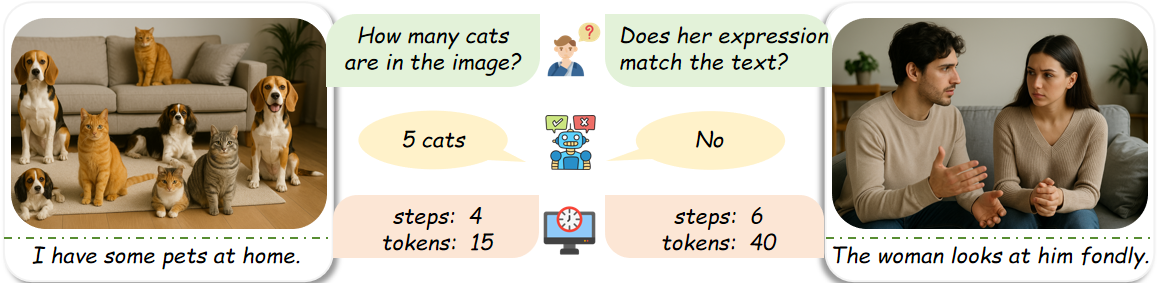}
   \caption{Reasoning complexity across multimodal tasks.}
   \label{fig:intro_1}
\vspace{- 0.5 cm}
\end{figure}

In the context of MLLMs, instance difficulty in the reasoning stage may exhibit substantial variation. Take \textbf{Figure~\ref{fig:intro_1}} as an example, examples of different difficulty levels require noticeably different numbers of generated tokens and reasoning steps, with harder instances exhibiting longer reasoning chains~\cite{wei2022chain,zhang2023multimodal,wang2025learning}. However, existing external reasoning methods typically apply fixed inference and sampling strategies to all instances, without dynamically adjusting to their varying difficulty~\cite{wang2022self,yang2023mm}. 
As a result, easy samples may often be over-computed, while complex ones receive insufficient computation. 
This imbalance may result in the issue compute-difficulty mismatch: uniform computation wastes resources on easy instances while failing to provide adequate coverage for hard ones, ultimately limiting efficiency and scalability.
To examine this, we conduct a motivation experiment on the MathVista benchmark \cite{lu2023mathvista} with both fixed and adaptive sampling strategies in \textbf{Section~\ref{sec:motivation_ex}}. 
The results show that fixed sampling may introduce significant redundancy, while difficulty-aware adaptive sampling maintains accuracy with fewer tokens. 
These findings highlight the necessity of an effective decoding method for adaptive computational allocation in MLLMs.

To explore and address the compute-difficulty mismatch issues, we begin by establishing a theoretical framework that connects sampling coverage, instance difficulty, and residual risk.
In this framework, the per-trial success probability is modeled as a random variable drawn from an instance-level difficulty distribution, capturing how heterogeneous multimodal samples contribute unequally to overall inference uncertainty.
The analysis shows that multimodal difficulty distributions are typically heavy-tailed: a small fraction of complex or ambiguous instances—such as those involving inconsistent visual–text alignment or deep reasoning dependencies—dominates the residual failure probability. This result reveals that fixed sampling strategies are inherently inefficient, as they allocate equal computation to instances of vastly different difficulty. 
The theory further implies that optimal efficiency is achieved when the sampling budget is adaptively distributed according to instance difficulty, ensuring that coverage converges at the fastest possible rate under a fixed compute constraint.

Building on this theoretical insight, we introduce a practical inference framework called Coverage-Aware Multimodal Decoding \textbf{(CAMD)}.
CAMD operationalizes the theoretical coverage–risk principles into an adaptive decoding mechanism that dynamically balances computational efficiency and inference reliability.
It integrates three components into a unified process: (i) evidence-weighted scoring to estimate single-sample success probability, (ii) posterior coverage estimation to assess whether current sampling achieves sufficient confidence, and (iii) sequential Bayesian updating to decide whether additional inference is warranted. By adaptively adjusting computation based on instance-level uncertainty, CAMD achieves a near-optimal trade-off between accuracy and efficiency, providing a theoretically grounded and practically effective solution to the compute–difficulty mismatch in multimodal large models.
Our contributions can be summarised as follows:
\begin{itemize}
    \item We establish a theoretical framework that characterizes the relationship between sampling coverage, instance difficulty, and residual risk in the reasoning of MLLMs;
    \item We propose Coverage-Aware Multimodal Decoding, CAMD, an adaptive inference framework that allocates computation based on instance-level uncertainty;
    \item Extensive experiments on multiple benchmarks validate the effectiveness and efficiency of the proposed method.
\end{itemize}

%% file: sec/2_rework.tex
\section{Related Work}
\label{re_work}

\subsection{Multimodal Large Language Models}
\label{sec:re_work_mllm}

Multimodal reasoning is a core capability of multimodal large language models (MLLMs)~\cite{Instructblip}, aiming to perform logical inference across modalities. Early work primarily focused on vision–language matching or visual question answering (VQA)~\cite{lu2019vilbert,yu2021ernie}, combining a visual encoder and a language model via shallow fusion. However, such approaches were limited in capturing deeper semantic and reasoning relationships. To address this, researchers have introduced Chain-of-Thought (CoT) techniques into vision–language settings, prompting models to generate intermediate ``reasoning chains" before final answers. For instance, the model LLaVA~\cite{liu2024llavanext} introduced by Liu uses visual instruction tuning on image-text instruction pairs to enable more general visual-language interaction. Further, system-level frameworks such as MM-REACT~\cite{yang2023mm} propose combining a large language model with specialized vision experts to carry out multimodal reasoning and action. Meanwhile, some recent efforts explore knowledge-graph grounding and multimodal knowledge graph fusion  to enhance reasoning robustness~\cite{marino2021krisp,chen2024knowledge}. In summary, the field has gradually moved from simple multimodal understanding toward structured, cross-modal logical inference and the generation of self-explanatory reasoning chains, laying the foundation for more advanced multimodal intelligence.

\subsection{Decoding Strategies for MLLMs}
\label{sec:re_work_de_rs}
Decoding strategies impact the reasoning performance of MLLMs. Classical methods like greedy and beam search often produce repetitive, low-diversity outputs \cite{holtzman2019curious, welleck2019neural}. Sampling-based methods, e.g., top-$k$ and nucleus sampling, improve diversity but struggle with coherence and factual consistency, especially at high temperatures \cite{fan2018hierarchical, kaplan2020scaling, zhu2024hot, jiang2024general}. Adaptive approaches, like min-$p$ and dynamic-temperature sampling, balance exploration and reliability but require fine-tuning. Recent advances, including Debiasing-Diversifying Decoding, reduce amplification bias at the cost of extra computation \cite{bao2024decoding}, while studies show that early token selection strongly affects downstream reasoning quality \cite{wei2024unveiling}. These findings highlight decoding as a key factor in both reasoning accuracy and diversity.
Beyond local sampling, scaling reasoning through multiple generation paths improves reliability. Best-of-$N$ decoding and self-consistency voting aggregate reasoning chains for better correctness \cite{wang2022self, snell2024scaling}, though their effectiveness decreases as computation grows, due to skewed difficulty distributions \cite{chowdhery2023palm}. To address this, planning-based and structured reasoning methods, such as Tree-of-Thoughts \cite{yao2023tree}, Monte Carlo Tree Search \cite{xie2024monte}, and Soft Reasoning \cite{xu2025phi}, expand exploration beyond linear decoding. While these methods show promise, they are computationally intensive in large-scale settings. This paper focuses on the trade-off between diversity, accuracy, and efficiency, aiming to dynamically allocate resources for more effective reasoning.

%% file: sec/3_analysis.tex
\section{Problem Settings and Analysis}

\subsection{Problem Settings}
\label{problem_set}

Given a dataset as $ \mathcal{D} = \{(x_v^i, x_t^i, y^i)\}_{i=1}^n$, where each sample consists of a visual input $x_v^i$, a textual input $x_t^i$, and the corresponding ground-truth answer $y^i$.  
For simplicity, we define a multimodal instance as $x_i = (x_v^i, x_t^i)$.
A MLLM first encodes the visual input $x_v^i$ into a sequence of visual tokens using a vision encoder (e.g., ViT), while the textual input $x_t^i$ is tokenized into a sequence of textual tokens through a language tokenizer. The visual and textual tokens are then concatenated and fed into a multimodal transformer backbone for autoregressive generation.  
Let the generated token sequence be $ o = \{o_1, o_2, \ldots, o_L\}$, where $L$ denotes the sequence length. The generation process follows the conditional probability distribution:
\begin{equation}
p_\theta(o \mid x_i) = \prod_{l=l}^{L} p_\theta(o_t \mid x_i, o_{<l}),
\end{equation}
where $\theta$ represents the model parameters, and $o_{<l}$ denotes all previously generated tokens.  
The sequence $o$ contains not only the final answer but also intermediate reasoning steps, reflecting the model’s chain-of-thought process.

In the inference stage, given a new input $x = (x_v, x_t)$, the model generates an output sequence through maximum-likelihood decoding: $o^* = \arg\max_o p_\theta(o \mid x)$, and the final predicted answer $y^*$ is extracted from $o^*$.
Conventional decoding strategies, such as Best-of-$N$ and Self-consistency decoding, typically adopt a fixed sampling budget across all samples. Given the intrinsic heterogeneity of multimodal reasoning tasks, this assumption is suboptimal—instance difficulty varies widely, and uniform computation results in inefficient coverage of the solution space. Consequently, adaptive sampling becomes essential for balancing inference reliability and computational efficiency.

\begin{figure*}[t]
    \centering
    \begin{subfigure}[t]{0.23\textwidth}
        \centering
        \includegraphics[width=\textwidth]{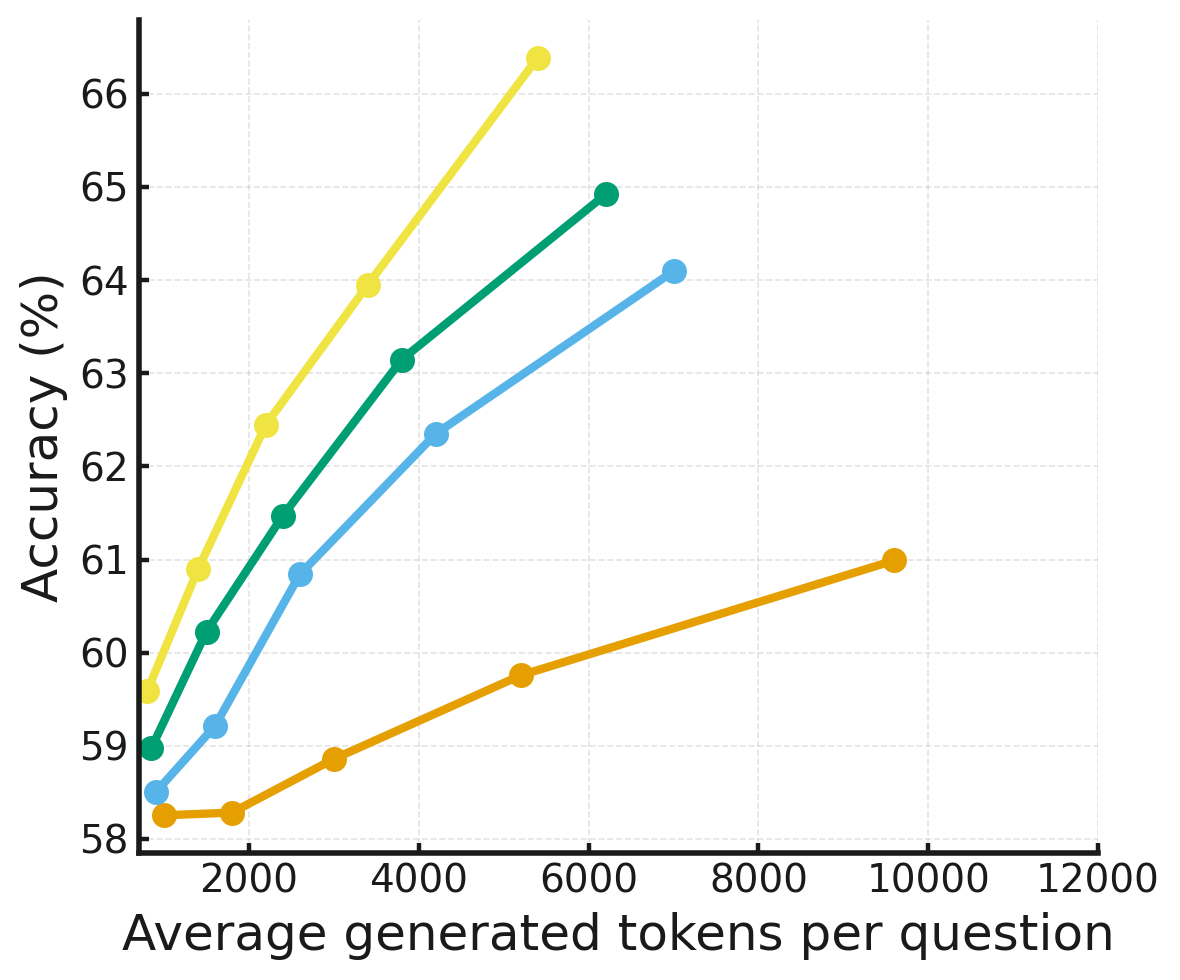}
        \caption{ACC vs. Tokens}
        \label{fig:effect1}
    \end{subfigure}
    \hfill
    \begin{subfigure}[t]{0.23\textwidth}
        \centering
        \includegraphics[width=\textwidth]{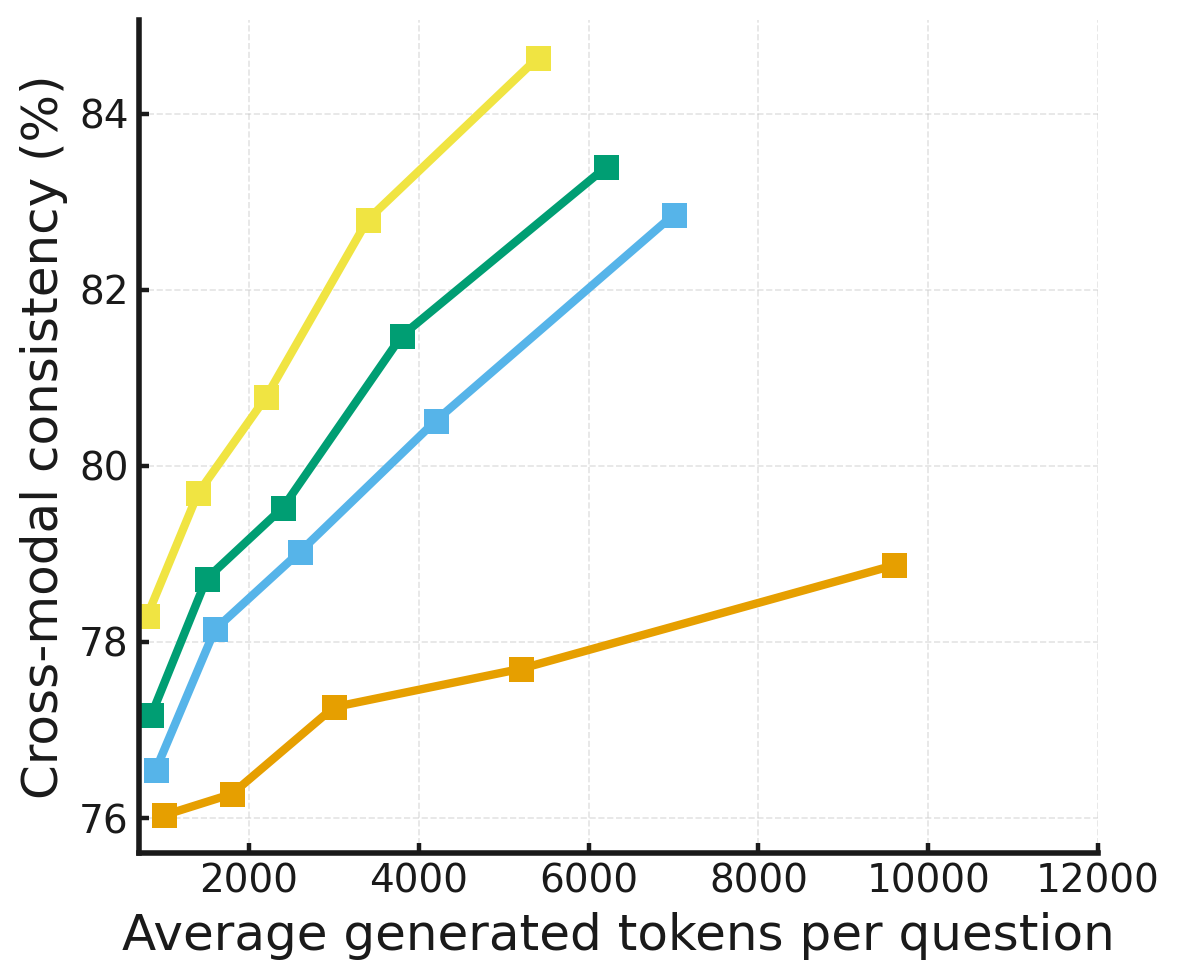}
        \caption{Consistency vs. Tokens}
        \label{fig:effect2}
    \end{subfigure}
    \hfill
    \begin{subfigure}[t]{0.23\textwidth}
        \centering
        \includegraphics[width=\textwidth]{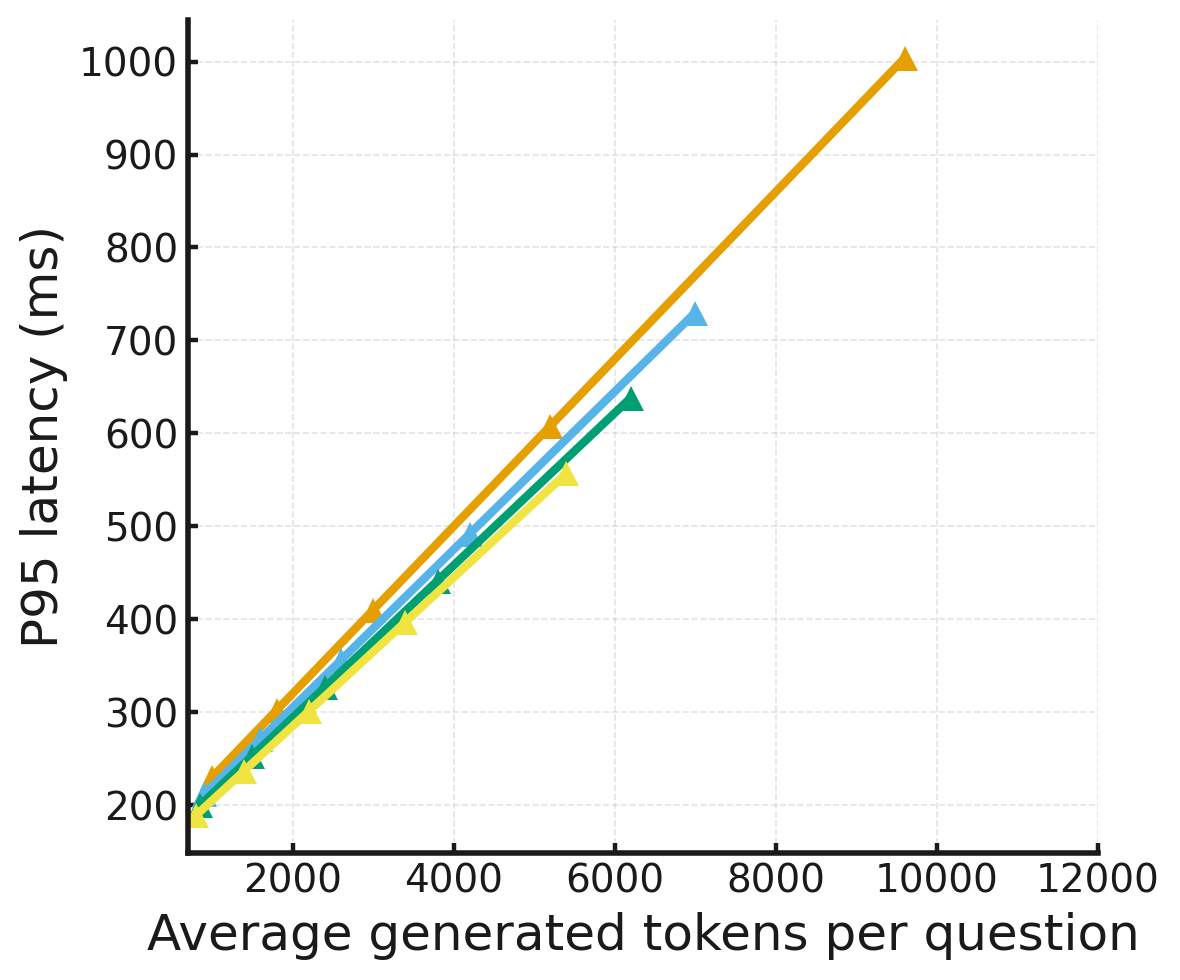}
        \caption{Latency vs Tokens}
        \label{fig:effect3}
    \end{subfigure}
    \hfill
    \begin{subfigure}[t]{0.23\textwidth}
        \centering
        \includegraphics[width=\textwidth]{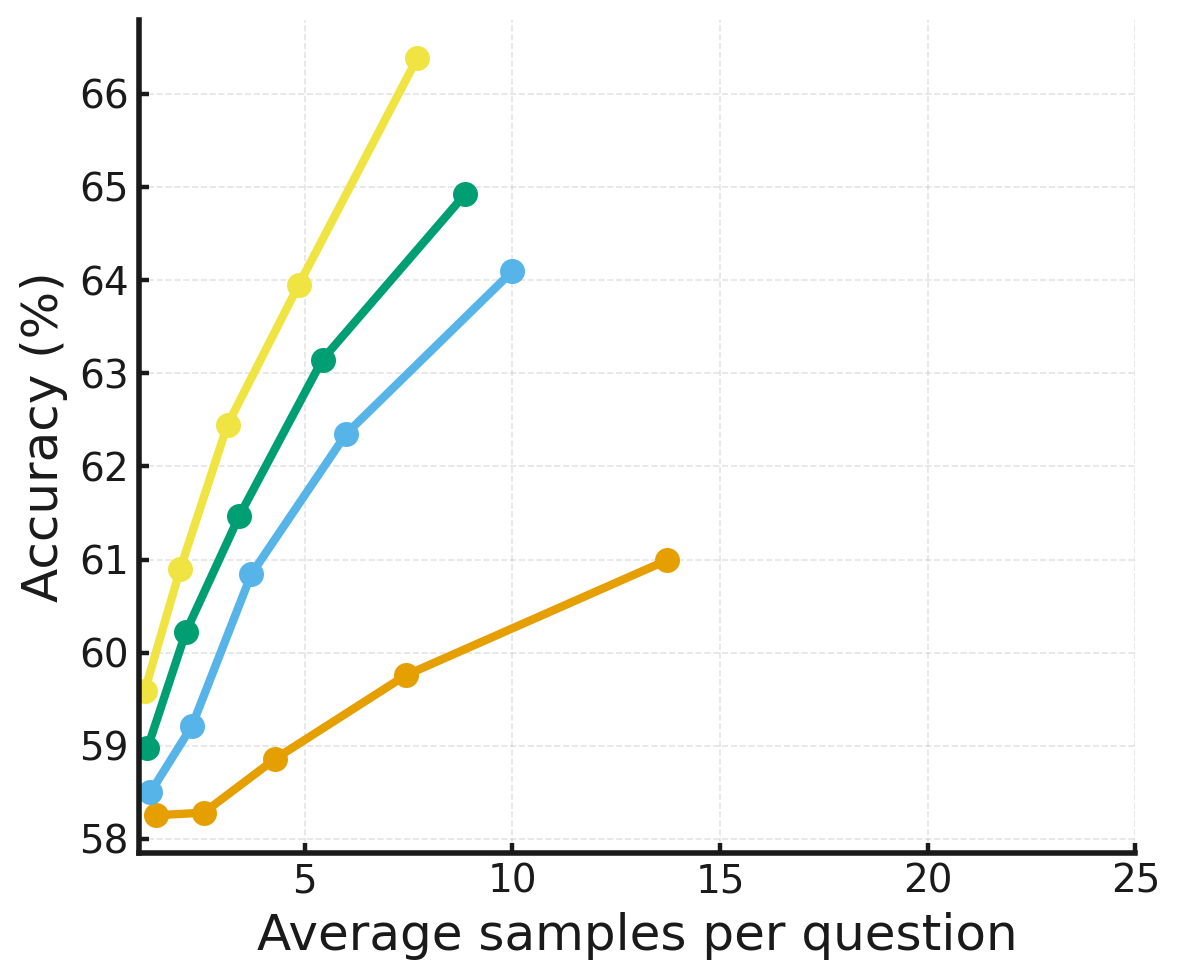}
        \caption{ACC vs. Avg Samples}
        \label{fig:effect4}
    \end{subfigure}
    \caption{Motivating experiments on MathVista. (a) Accuracy (\%) vs. average generated tokens; (b) Cross-modal consistency (\%) vs. tokens; (c) P95 latency (ms) vs. tokens; (d) Accuracy (\%) vs. average samples. }
    \label{fig:motivation}
\end{figure*}

\subsection{Emprical Analysis}
\label{sec:motivation_ex}
To evaluate the effect of decoding strategies, we conduct a toy experiment on MLLMs.
We evaluate Qwen2-VL-7B-Instruct on the MathVista benchmark \cite{lu2023mathvista}, which consists of 6,141 visual-mathematical reasoning problems across diverse domains. Each question contains both visual and textual components, and answers are normalized through symbolic or numerical equivalence checking to ensure objective evaluation.
Visual features are extracted once per image and cached for all subsequent samples, ensuring that increased sampling only affects the language decoding stage. Decoding follows a stochastic setup with temperature $T=0.7$, top-p = 0.9, repetition penalty = 1.05, and a maximum generation length of 16,384 tokens. We adopt best-of-N decoding as the fixed baseline ($N\in \{1,2,4,8,16,32 \}$) and use $(N=256)$ as an approximate upper bound to represent complete sampling coverage. In addition, we implement three adaptive stopping strategies: (i) a threshold-based rule that halts generation once a satisfactory answer score is reached or when no improvement occurs over three consecutive samples; (ii) a Bayesian Beta–Bernoulli posterior update that terminates when the expected success probability fails to justify further sampling; and (iii) an Expected Improvement (EI) criterion that stops when the estimated marginal gain in answer quality falls below the per-token computational cost. To ensure fair comparison, each adaptive rule determines its sampling budget online using model-derived difficulty proxies, i.e., the log-probability confidence, rather than ground-truth difficulty levels. Semantic deduplication cosine similarity $> 0.9$ is applied to remove redundant generations.

The results are shown in \textbf{Figure \ref{fig:motivation}}.
We observe that multi-sampling in MLLM inference is not computationally free. The total number of generated tokens and wall-clock latency increase almost linearly with the sampling count, while accuracy and cross-modal consistency quickly saturate after moderate sampling (typically $N\ge8$), revealing strong diminishing returns. Fixed-$N$ decoding leads to substantial oversampling on simpler visual-textual questions, especially in chart or geometric recognition tasks, where additional samples rarely alter correctness but add significant computational overhead. In contrast, adaptive stopping dynamically tailors computation to the instance difficulty: on easier problems, the average sampling number drops to roughly $2-3$ without any loss in accuracy compared with fixed $N=8$; on more complex visual reasoning cases, adaptive rules automatically expand the sampling budget to $32-64$, achieving accuracy close to the $N=256$ upper bound at only a fraction of the computational cost. Notably, cross-modal consistency verification effectively guides adaptive sampling toward ambiguous or conflicting visual contexts, reducing hallucination-related errors without increasing total inference cost.

Therefore, the commonly used fixed-$N$ decoding is not compute-optimal for MLLMs. Compared to unimodal LLM settings, the benefits of adaptive sampling in MLLMs are more pronounced, as visual–textual grounding introduces additional complexity and hallucination risk. This inspires us to allocate computational resources adaptively to harder or ambiguous visual inputs, aiming to establish a new Pareto frontier between reasoning accuracy and efficiency.

%% file: sec/4_methods.tex
\section{Methodology}
\label{sec:Methodology}

In this section, we provide a comprehensive analysis of the compute–difficulty mismatch in the reasoning of MLLMs. We first establish a theoretical framework that formalizes the relationship among sampling size, coverage probability, and instance difficulty. Based on the theoretical results, we then propose CAMD, a Coverage-Aware Multimodal Decoding method that adaptively allocates sampling by instance difficulty and evidence-weighted scoring to ensure both effectiveness and efficiency.

\subsection{Theoretical Foundations}
\label{subsec:theory}

In the reasoning stage of MLLMs, models must fuse information from multiple modalities. Instance difficulty varies widely: cross-modal inconsistencies or ambiguities often demand more computation to produce reliable answers. To characterize this variability in a modality-agnostic way, we introduce a simple theoretical framework that links instance difficulty to computational demand.

Specifically, we first formalize the single-trial success probability and the $K$-trial coverage as basic metrics that relate sampling budget to performance. For each instance $x=(x_v,x_t)$, let $s \in(0,1)$ denote the probability that a single inference trial produces the correct result. 
Because instances vary in intrinsic difficulty, $s$ is treated as a random variable drawn from an instance-level distribution $G(s)$ with density $g(s)$.  In MLLMs, $s$ reflects the model’s per-trial success probability conditioned on modality consistency and reasoning depth.
When performing $K$ independent trials for the same input, the probability of obtaining at least one correct result, termed the \emph{coverage}, is:
\begin{equation}
\mathcal{C}(K) = \mathbb{E}_{s\sim G}\!\left[1-(1-s)^K\right],
\end{equation}
and the expected residual failure probability is
\begin{equation}
\Delta(K) = 1 - \mathcal{C}(K) = \mathbb{E}_{s\sim G}\!\left[(1-s)^K\right].
\end{equation}
These two quantities jointly characterize how performance scales with the sampling budget, where $\mathcal{C}(K)$ denotes expected coverage and $\Delta(K)$ the residual failure rate.

To quantify the number of samples required for a desired confidence level, we define the notion of $\delta$-coverage.
\begin{definition}[$\delta$-coverage sampling size $N_\delta$]\label{defini:1}
Given $x$ with single-trial success probability $s$, the minimal number of samples required to reach confidence level $1-\delta$ is:
\begin{equation}
N_\delta=\min\{n:1-(1-s)^n\ge 1-\delta\}= \frac{\log\delta}{\log(1-s)},
\end{equation}
drawing $N_\delta$ independent samples guarantees that the probability of obtaining at least one correct answer at least $1-\delta$.
\end{definition}
\textbf{Definition \ref{defini:1}} links sample efficiency to instance difficulty: smaller $s$ require substantially more samples to reach the same confidence, i.e., for $s\ll1$ one has $N_{\delta}\approx -\log\delta / s$, where the sample demand grows roughly like $1/s$.

Because difficult instances correspond to small $s$, the decay of residual failure $\Delta(K)$ is governed by the lower tail of $G(s)$. Next, we establish how the shape of this tail determines the convergence rate:
\begin{theorem}[Tail-dominated convergence rate]
Let $g(s)$ denote the density of $G(s)$ near $s\to0$. Then the asymptotic decay of $\Delta(K)$ follows:
\begin{itemize}
    \item \textbf{Heavy tail (polynomial):} if $g(s)\!\sim\!\kappa s^{\alpha-1}\ell(1/s)$ with $\ell$ slowly varying, then
    \[
    \Delta(K)\sim \kappa\Gamma(\alpha)K^{-\alpha}\ell(K).
    \]
    \item \textbf{Stretched-exponential tail:} if $\log\Pr(s\!\le\!\varepsilon)\!\sim\!-c\varepsilon^{-\theta}$, then
    \[
    \log\Delta(K)\sim -C_\theta K^{\theta/(\theta+1)}.
    \]
    \item \textbf{Light or truncated tail:} if $G([0,\varepsilon])\!\le\! C e^{-c_0/\varepsilon}$, then
    \[
    \Delta(K)\!\le\!C'e^{-c'K}.
    \]
\end{itemize}
\label{thm_tdcr}
\end{theorem}
\textbf{Theorem \ref{thm_tdcr}} shows that intuitively, heavier tails imply that a substantial fraction of instances have extremely small success probability, leading to slow (power-law) convergence; lighter tails enable exponential improvement with $K$. 

While the above theorem characterizes how the expected residual failure $\Delta(K)$ decays with sampling size, in practice, the total risk may include an irreducible component that cannot be reduced by further sampling. 
Further, we decompose the overall failure probability $\mathcal{R}(K)$ as:
\begin{equation}
\mathcal{R}(K)=\Delta(K)+\mathcal{R}_{\mathrm{irr}},
\end{equation}
where $\mathcal{R}_{\mathrm{irr}}$ denotes irreducible risk arising from unresolvable perceptual or definitional uncertainty (e.g., occluded images or ill-posed queries). 
Let $\epsilon \in (0,1)$ denote a target upper bound on the overall failure probability—that is, the desired maximum tolerable risk level for inference.
To achieve $\mathcal{R}(K)\le\epsilon$, it suffices that $\Delta(K)\le \epsilon-\mathcal{R}_{\mathrm{irr}}$.

Combining this with Theorem~\ref{thm_tdcr} yields the asymptotic scaling of the minimal required sampling size:
\begin{equation}
\begin{aligned}
K^\star(\epsilon) & \asymp
&
\begin{cases}
\bigl({\kappa\Gamma(\alpha)}/{(\epsilon-\mathcal{R}_{\mathrm{irr}})}\bigr)^{1/\alpha}, & \text{heavy tail},\\[4pt]
\log^{(\theta+1)/\theta}\!\!\bigl(1/(\epsilon-\mathcal{R}_{\mathrm{irr}})\bigr), & \text{stretched-exp},\\[4pt]
\log\!\!\bigl(1/(\epsilon-\mathcal{R}_{\mathrm{irr}})\bigr), & \text{light tail}.
\end{cases}
\end{aligned}
\end{equation}

The above analysis reveals that the efficiency of multimodal inference is fundamentally limited by the tail of the instance-difficulty distribution. 
Under heavy-tailed difficulty, additional samples yield diminishing returns, while lighter tails enable rapid risk reduction. 
Therefore, fixed sampling budgets are inherently suboptimal-adaptive compute allocation based on estimated instance difficulty is essential for optimal efficiency.

\begin{figure*}[t]
    \centering
\includegraphics[width=\linewidth]{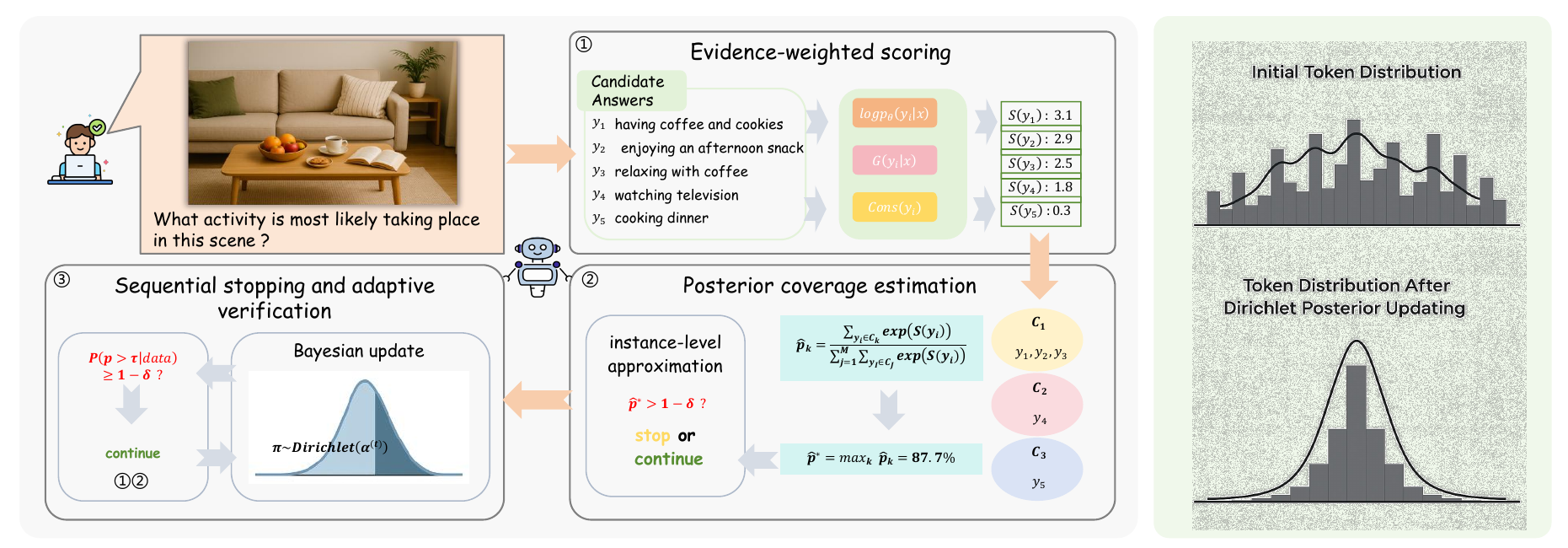}
    \caption{The framework of the proposed CAMD (Coverage-Aware Multimodal Decoding) for MLLMs.}
    \label{fig:placeholder}
\end{figure*}

\subsection{Coverage-Aware Multimodal Decoding}
\label{subsec:method}

Building upon the above discussion, we propose Coverage-Aware Multimodal Decoding (CAMD), a decoding framework that operationalizes the relationship between sampling coverage and residual risk for more effective reasoning. 
CAMD operationalizes the theoretical coverage–risk relationship through three key modules:  
(1) \textit{evidence-weighted scoring}, quantifying candidate-level confidence and multimodal consistency to approximate the single-trial success probability $s$;  
(2) \textit{posterior coverage estimation}, aggregating these scores into an empirical posterior to assess whether the current sample set sufficiently covers the correct answer; and  
(3) \textit{Bayesian adaptive sampling }, performing Bayesian posterior updates and dynamically reweights token probabilities to allocate computation according to instance-level uncertainty.

\subsubsection{Evidence-weighted scoring}
\label{mehtod_sub1}

Following the theoretical formulation where the per-trial success probability $s$ governs coverage, we approximate $s$ empirically using an evidence-weighted score. Given a multimodal input $x=(x_v,x_t)$, the model generates candidate answers $\{y_i\}_{i=1}^{K_t}$ at each sampling step, where $K_t$ denotes the cumulative number of samples, which increases dynamically until the coverage criterion is satisfied. Here, we design an evidence-weighted scoring function that integrates three components:(1) generation confidence, (2) cross-modal consistency, and (3) reasoning coherence.

\textbf{Generation confidence.} For each candidate answer $y_i = \{y_{i,1}, y_{i,2}, \dots, y_{i,L_i}\}$,  where $L_i$ denotes the token length, we measure the model’s intrinsic certainty through the normalized log-likelihood of each token:
\begin{equation}
S_{\text{gen}}(y_i|x)=\frac{1}{L_i}\sum_{t=1}^{L_i}\log p_\theta(y_{i,t}|y_{i,<t},x).
\end{equation}
This reflects how confidently the model generates the sequence under the given multimodal evidence,  
thereby capturing the likelihood of success for a single decoding trial.

\textbf{Cross-modal consistency.} To ensure that generated tokens remain semantically aligned with the visual–text inputs, we compute token-level alignment between embeddings of generated tokens and multimodal evidence features.  Let $\mathcal{E}_v(x_v)=\{v_j\}_{j=1}^{N_v}$ and $\mathcal{E}_t(x_t)=\{t_r\}_{r=1}^{N_t}$  
denote visual and textual evidence features, with encoders $f_v(\cdot)$ and $f_t(\cdot)$, respectively.  
The token-level consistency is:
\begin{equation}
\begin{aligned}
    G(y_{i,t}|x)= & \tfrac{1}{2}\!
\frac{1}{N_v}\sum_{j=1}^{N_v}\cos(f_v(v_j),f_t(y_{i,t})) \\
& + \frac{1}{N_t}\sum_{r=1}^{N_t}\max_{1\le j\le N_v}\cos(f_t(t_r),f_v(v_j)),
\end{aligned}
\end{equation}
and the overall cross-modal alignment score is:
\begin{equation}
S_{\text{align}}(y_i|x)=\frac{1}{L_i}\sum_{t=1}^{L_i}G(y_{i,t}|x).
\end{equation}
Higher alignment indicates stronger grounding between visual and textual evidence,  
therefore increasing the empirical probability that the generated output contributes to the true coverage of the correct answer, i.e., raising the expected $s$.

\textbf{Reasoning coherence.}
Reasoning coherence measures the logical smoothness of the generated reasoning process. For each pair of consecutive tokens, we compute similarity between their hidden or embedding representations:
\begin{equation}
Cons(y_{i,t})=\cos(h_{i,t},h_{i,t+1}),
\end{equation}
and aggregate as:
\begin{equation}
S_{\text{coh}}(y_i)=\frac{1}{L_i-1}\sum_{t=1}^{L_i-1}Cons(y_{i,t}).
\end{equation}
When hidden states are inaccessible, $h_{i,t}$ is replaced by the embedding $f_t(y_{i,t})$ to approximate semantic continuity.  A higher $S_{\text{coh}}$ implies smoother reasoning transitions, reducing local inconsistencies and improving reasoning stability.

Integrating the above components, the evidence-weighted scoring function is:
\begin{equation}
S(y_i|x)=
S_{\text{gen}}(y_i|x)
+\lambda_g\,S_{\text{align}}(y_i|x)
+\lambda_c\,S_{\text{coh}}(y_i),
\label{equ:S_final}
\end{equation}
where $\lambda_g$ and $\lambda_c$ are weighting coefficients. The normalized score $\tilde{s}_i=\mathrm{softmax}(S(y_i|x))$
provides an empirical approximation of the per-trial success probability $s_i$, which serves as the foundation for posterior coverage estimation in the subsequent module.  This formulation operationalizes the theoretical linkage between coverage and per-trial success, allowing CAMD to estimate coverage probability directly from multimodal evidence.

\subsubsection{Posterior coverage estimation}
\label{mehtod_sub2}

According to Definition \ref{defini:1},  sampling can stop once the probability of including at least one correct answer exceeds a target confidence level $1-\delta$. To operationalize this principle in practice, we estimate the coverage posterior from the generated candidates by grouping semantically similar answers into clusters.  A large language model (LLM) is employed to compute pairwise semantic similarities $\mathrm{Sim}_{ij}$ considering both textual and multimodal evidence,  
and to dynamically determine the cluster assignments:
\begin{equation}
\{C_k\}_{k=1}^{M_t} = \mathrm{Cluster}_{\text{LLM}}\!\left(\{y_i\}_{i=1}^{K_t}, \{\mathrm{Sim}_{ij}\}\right),
\end{equation}
where $M_t$ denotes the number of clusters adaptively determined at iteration $t$. Within each semantic cluster $C_k$, we compute a normalized posterior weight based on the aggregated evidence-weighted scores:
\begin{equation}
\hat{p}_k=\frac{\sum_{y_i\in C_k}\exp(S(y_i|x))}{\sum_{j=1}^{M}\sum_{y_i\in C_j}\exp(S(y_i|x))},
\end{equation}
where $S(y_i|x)$ is the evidence-weighted score. The maximal posterior coverage estimate is $\hat{p}^\star=\max_k \hat{p}_k$,
which quantifies the probability that the most probable semantic cluster already contains at least one correct answer.

If $\hat{p}^\star \ge 1-\delta$, the current sampling set is considered sufficient and decoding terminates early. Otherwise, additional evidence refinement is triggered. This criterion serves as a practical approximation to the theoretical $\delta$-coverage condition, achieving adaptive stopping without requiring ground-truth labels. If the estimated coverage $\hat{p}^\star$ is below the confidence threshold $(1-\delta)$, the decoding process proceeds to the next stage, where adaptive token sampling is performed to refine evidence and improve coverage estimation. 

\subsubsection{Bayesian adaptive sampling}
\label{mehtod_sub3}

As discussed in Section~\ref{subsec:theory}, the single-trial success probability $s$ governs coverage by describing the chance that a single decoding attempt includes the correct answer. In practice, however, a multimodal instance often gives rise to multiple semantic hypotheses (clusters), each representing a distinct reasoning path with its own probability of being correct. To extend the theoretical formulation to this multi-hypothesis setting and enable adaptive token weighting in the next sampling round, we replace the scalar success probability $s$ with a vector of cluster-level probabilities, i.e., $\boldsymbol{\pi}^{(t)} = (\pi_1^{(t)}, \ldots, \pi_{M_t}^{(t)})$,
where $\pi_k^{(t)}=\Pr(C_k\text{ contains at least one correct answer})$, and $M_t$ denotes the number of clusters in iteration $t$.

This vectorized formulation jointly models uncertainty across semantic clusters and serves as the basis for adaptive token reweighting in subsequent decoding. We adopt a Dirichlet prior for $\pi$, i.e., $\pi \sim \mathrm{Dirichlet}(\boldsymbol{\alpha}^{(t)})$,
which serves as the conjugate prior of a multinomial model and provides a natural bridge between the theoretical difficulty distribution $G(s)$ and the learnable cluster-wise success probabilities. When $\hat{p}^\star < 1-\delta$, adaptive sampling is triggered. Given the current evidence strength within each cluster, represented as soft counts $n_k^{(t)}=\sum_{y_i\in C_k}\tilde{s}_i$,
the posterior of $\pi$ is updated via Bayesian inference:
\begin{equation}
\begin{aligned}
    & \pi \mid \mathcal{D}_t \sim 
\mathrm{Dirichlet}\!\big(\boldsymbol{\alpha}^{(t)}+\mathbf{n}^{(t)}\big),\\
    &\bar{\pi}_k^{(t)}=\mathbb{E}[\pi_k\mid \mathcal{D}_t]
=\frac{\alpha_k^{(t)}+n_k^{(t)}}{\sum_j(\alpha_j^{(t)}+n_j^{(t)})}.
\end{aligned}
\end{equation}
where $\mathcal{D}_t$ denotes all candidate evidence observed up to iteration $t$. Then, we use the expected cluster weights $\bar{\pi}_k^{(t)}$ as adaptive sampling coefficients to guide the next-round decoding. Specifically, we define a mixture token distribution that integrates information across all clusters:
\begin{equation}
p'(y_t) = \sum_{k=1}^{M_t} \bar{\pi}_k^{(t)}\, q_k(y_t),
\end{equation}
where $q_k(y_t)$ denotes the token-level conditional distribution within cluster $C_k$, following the evidence-weighted formulation in Eq.~\ref{equ:S_final}.  This mixture distribution adaptively reweights token probabilities according to the posterior confidence of each semantic cluster, allowing the model to focus sampling on more promising reasoning directions while maintaining global diversity.

In summary, CAMD operationalizes the theoretical coverage-risk framework into a practical decoding process. By iteratively updating posterior cluster weights and reweighting token probabilities, it dynamically allocates computation according to instance-level uncertainty, ensuring both reasoning accuracy and efficiency.

%% file: sec/5_exp.tex
\begin{table*}[t]
    \centering
        \caption{Comparison across image benchmarks. $\uparrow$ indicates higher-is-better, while $\downarrow$ indicates lower-is-better. The values in parentheses report the improvement of CAMD over the base model. The best results are highlighted in \textbf{bold}. More results are shown in \textbf{Appendix E}.}
    \vspace{-0.15cm}
    \footnotesize
    \resizebox{\textwidth}{!}{
    \begin{tabular}{l| c c c | c c | c c c c c}
         \hline
        \rowcolor{gray!10}
            \multicolumn{1}{l|}{\cellcolor{gray!10}} & \multicolumn{3}{c|}{\cellcolor{gray!10}\textbf{Comprehensive Benchmark}} & \multicolumn{2}{c|}{\cellcolor{gray!10}\textbf{General VQA}} & \multicolumn{5}{c}{\cellcolor{gray!10}\textbf{Hallucination Benchmark}} \\
        \cline{2-11}
        \rowcolor{gray!10}
\multicolumn{1}{l|}{\multirow{-2}{*}{\cellcolor{gray!10}\textbf{Method}}} & \textbf{MMBench $\uparrow$} & \textbf{LLaVA$^{\mathrm{W}}$} & \textbf{MM-Vet$\uparrow$} & \textbf{VizWiz$\uparrow$} & \textbf{SQA$\uparrow$} & \textbf{CHAIR$_{S}$ $\downarrow$} & \textbf{CHAIR$_{I}$ $\downarrow$} & \textbf{POPE-R$\uparrow$} & \textbf{POPE-P$\uparrow$} & \textbf{POPE-A$\uparrow$} \\
        \hline
        LLaVA-1.5 & 64.3 & 72.5 & 30.5 & 48.5 & 64.5 & 48.0 & 13.9 & 87.0 & 82.8 & 76.6 \\
        +ICD & 63.1 & 69.7 & 30.4 & 46.9 & 62.8 & 47.7 & 13.6 & 87.9 & 84.0 & 80.2 \\
        +VCD & 63.9 & 70.9 & 29.5 & 43.4 & 63.3 & 46.8 & 13.2 & 87.0 & 83.5 & 78.1 \\
        +CGD & 64.5 & 71.3 & 30.8 & 49.0 & 64.0 & 46.2 & 13.5 & 87.5 & 83.8 & 79.0 \\
        +OPERA & 64.4 & 72.0 & 31.4 & 50.0 & 64.9 & 45.2 & 12.7 & 88.8 & 82.8 & 79.2\\
        + FarSight & 66.0 & 74.7 & 32.5 & 50.8 & 67.4 &  41.6 & 13.2 & 90.5 & 86.1 & 80.4 \\
        \rowcolor{orange!10} \textbf{+ CAMD (Ours)} & \textbf{66.6~\color{blue!60}{(+2.3)}} & \textbf{75.0~\color{blue!60}{(+2.5)}} & \textbf{33.1~\color{blue!60}{(+2.6)}} & \textbf{51.2~\color{blue!60}{(+2.7)}} & \textbf{68.0~\color{blue!60}{(+3.5)}} & \textbf{40.6~\color{blue!60}{(-7.4)}} & \textbf{12.9~\color{blue!60}{(-1.0)}} & \textbf{91.0~\color{blue!60}{(+4.0)}} & \textbf{86.5~\color{blue!60}{(+3.7)}} & \textbf{81.2~\color{blue!60}{(+4.6)}} \\
        \cdashline{1-11}
        InstructBLIP & 43.4 & 58.2 & 25.6 & 33.4 & 62.1 & 55.6 & 24.2 & 88.7 & 81.3 & 74.4 \\
        + FarSight & 46.5 & 61.0 & 27.8 & 36.0 & 63.4 & 51.8 & 23.0 & 89.5 & 85.8 & 76.7 \\
        \rowcolor{orange!10} \textbf{+ CAMD (Ours)} & \textbf{46.8~\color{blue!60}{(+3.4)}} & \textbf{61.2~\color{blue!60}{(+3.0)}} & \textbf{28.2~\color{blue!60}{(+2.6)}} & \textbf{36.6~\color{blue!60}{(+3.2)}} & \textbf{63.6~\color{blue!60}{(+1.5)}} & \textbf{51.2~\color{blue!60}{(-4.4)}} & \textbf{22.5~\color{blue!60}{(-1.7)}} & \textbf{89.6~\color{blue!60}{(+1.1)}} & \textbf{86.1~\color{blue!60}{(+4.8)}} & \textbf{77.0~\color{blue!60}{(+2.6)}} \\
\cdashline{1-11}

        Video-LLaVA  & 60.9 & 73.1 & 32.0 & 48.1 & 64.6 & 50.2 & 15.6 & 81.6 & 85.3 & 86.2 \\
        + FarSight & 62.8 & 74.5 & 32.8 & 50.3 & 66.2 & 44.8 & 12.9 & 83.2 & 85.8 & 87.1 \\
        
        \rowcolor{orange!10} \textbf{+ CAMD (Ours)} & \textbf{63.1~\color{blue!60}{(+2.2)}} & \textbf{75.0~\color{blue!60}{(+1.9)}} & \textbf{33.3~\color{blue!60}{(+1.3)}} & \textbf{50.8~\color{blue!60}{(+2.7)}} & \textbf{66.9~\color{blue!60}{(+2.3)}} & \textbf{44.3~\color{blue!60}{(-5.9)}} & \textbf{12.1~\color{blue!60}{(-3.5)}} & \textbf{84.2~\color{blue!60}{(+2.6)}} & \textbf{86.2~\color{blue!60}{(+0.9)}} & \textbf{87.8~\color{blue!60}{(+1.6)}} \\
\cdashline{1-11}

        Chat-UniVi & 56.3 & 70.4 & 28.3 & 46.9 & 59.9 & 52.3 &  16.7 & 85.1 & 69.5 & 64.4 \\
        + FarSight & 59.8 & 72.6 & 30.7 & 48.2 & 62.4 & 48.9 & 15.2 & 87.4 & 69.7 & 65.3 \\
        
        \rowcolor{orange!10} \textbf{+ CAMD (Ours)} & \textbf{60.6~\color{blue!60}{(+4.3)}} & \textbf{73.6~\color{blue!60}{(+3.2)}} & \textbf{31.4~\color{blue!60}{(+3.1)}} & \textbf{48.8~\color{blue!60}{(+1.9)}} & \textbf{63.0~\color{blue!60}{(+3.1)}} & \textbf{48.2~\color{blue!60}{(-4.2)}} & \textbf{14.7~\color{blue!60}{(-2.0)}} & \textbf{87.9~\color{blue!60}{(+2.8)}} & \textbf{70.3~\color{blue!60}{(+0.8)}} & \textbf{65.9~\color{blue!60}{(+1.5)}} \\
        \bottomrule
    \end{tabular}}
    \vspace{-0.1cm}
    \label{tab:ex_image_result}
\end{table*}

\begin{table*}[t]
\centering
\caption{Comparison across all video benchmarks. For the Video-Based Text Generation benchmark, we evaluate five dimensions following \cite{tang2025seeing}: \textbf{Cr.} (Correctness of Information), \textbf{Cs.} (Consistency), \textbf{De.} (Detail Orientation), \textbf{Ct.} (Contextual Understanding), and \textbf{Te.} (Temporal Understanding). Following~\cite{Maaz2023VideoChatGPT}, we use GPT-3.5 Turbo to assign relative scores to model outputs on a 0-5 scale.}
\vspace{-0.2cm}
\small
\setlength{\tabcolsep}{10pt}
\resizebox{\linewidth}{!}{
\begin{tabular}{l| c c| c c| c c c c c}
\hline
\rowcolor{gray!10}
\multicolumn{1}{l|}{\cellcolor{gray!10}} & \multicolumn{2}{c|}{\textbf{MSVD-QA}} & \multicolumn{2}{c|}{\textbf{ActivityNet-QA}} & \multicolumn{5}{c}{\textbf{Video-Based Text Generation}} \\
\cline{2-10}
\rowcolor{gray!10}
\multicolumn{1}{l|}{\multirow{-2}{*}{\cellcolor{gray!10}\textbf{Method}}} & \textbf{Accuracy$\uparrow$} & \textbf{Score$\uparrow$} & \textbf{Accuracy$\uparrow$} & \textbf{Score$\uparrow$} & \textbf{Cr.$\uparrow$} & \textbf{Cs.$\uparrow$} & \textbf{De.$\uparrow$} & \textbf{Ct.$\uparrow$} & \textbf{Te.$\uparrow$} \\
 \hline
 
Chat-UniVi  & 64.6 & 3.6 & 43.1 & 3.2 & 2.84 & 2.93 & 2.55 & 3.16 & 2.43 \\
+ FarSight & 66.4  & 3.5 & 43.7 & 3.2 & 2.86 & 2.94 & 2.56 & 3.19 & 2.48 \\
\rowcolor{orange!10}\textbf{+ CAMD (Ours)} & \textbf{67.0~\color{blue!60}{(+2.4)}}  & \textbf{3.8~\color{blue!60}{(+0.2)}} & \textbf{44.2~\color{blue!60}{(+1.1)}} & \textbf{3.3~\color{blue!60}{(+0.1)}} & \textbf{3.05~\color{blue!60}{(+0.21)}} & \textbf{3.10~\color{blue!60}{(+0.17)}} & \textbf{2.72~\color{blue!60}{(+0.17)}} & \textbf{3.35~\color{blue!60}{(+0.19)}} & \textbf{2.60~\color{blue!60}{(+0.17)}} \\

Video-LLaVA  & 64.8 & 3.7 & 41.5 & 3.3 & 2.32 & 2.34 & 2.65 & 2.75 & 2.09 \\
+ FarSight & 66.2 & 3.6 & 42.0 & 3.5 & 2.43 & 2.38 & 2.93 & 2.84 & 2.14 \\
\rowcolor{orange!10}\textbf{+ CAMD (Ours)} & \textbf{66.9~\color{blue!60}{(+2.1)}} & \textbf{3.9~\color{blue!60}{(+0.2)}} & \textbf{42.3~\color{blue!60}{(+0.8)}} & \textbf{3.6~\color{blue!60}{(+0.3)}} & \textbf{2.58~\color{blue!60}{(+0.26)}} & \textbf{2.55~\color{blue!60}{(+0.21)}} & \textbf{3.06~\color{blue!60}{(+0.41)}} & \textbf{3.02~\color{blue!60}{(+0.27)}} & \textbf{2.30~\color{blue!60}{(+0.21)}} \\

VILA & 72.6 & 4.0 & 50.2 & 3.3 & 3.14 & 3.40 & 2.71 & 3.43 & 2.58 \\
+ FarSight & 74.5 & 4.2 & 51.4 & 3.6 & 3.18 & 3.52 & 2.73 & 3.45 & 2.60 \\
\rowcolor{orange!10}\textbf{+ CAMD (Ours)} & \textbf{75.1~\color{blue!60}{(+2.5)}} & \textbf{4.3~\color{blue!60}{(+0.3)}} & \textbf{52.0~\color{blue!60}{(+1.8)}} & \textbf{3.8~\color{blue!60}{(+0.5)}} & \textbf{3.30~\color{blue!60}{(+0.16)}} & \textbf{3.65~\color{blue!60}{(+0.25)}} & \textbf{2.95~\color{blue!60}{(+0.24)}} & \textbf{3.72~\color{blue!60}{(+0.29)}} & \textbf{2.85~\color{blue!60}{(+0.27)}} \\

Video-LLaMA2   & 70.9 &  3.8  & 49.9 & 3.3 & 3.13 & 3.23 & 2.70 & 3.42 & 2.45 \\
+ FarSight & 73.8 & 3.9 & 50.4 & 3.6 & 3.26 &  3.32 & 3.21 & 3.50 & 2.47 \\
\rowcolor{orange!10}\textbf{+ CAMD (Ours)} & \textbf{74.6~\color{blue!60}{(+3.7)}} & \textbf{4.0~\color{blue!60}{(+0.2)}} & \textbf{50.9~\color{blue!60}{(+1.0)}} & \textbf{3.7~\color{blue!60}{(+0.4)}} & \textbf{3.51~\color{blue!60}{(+0.38)}} & \textbf{3.57~\color{blue!60}{(+0.34)}} & \textbf{3.45~\color{blue!60}{(+0.75)}} & \textbf{3.73~\color{blue!60}{(+0.31)}} & \textbf{2.62~\color{blue!60}{(+0.17)}} \\
\bottomrule
\end{tabular}}
\label{tab:ex_video_result}
\vspace{-0.2cm} 
\end{table*}

\begin{figure}[t]
    \centering
    \includegraphics[width=\linewidth]{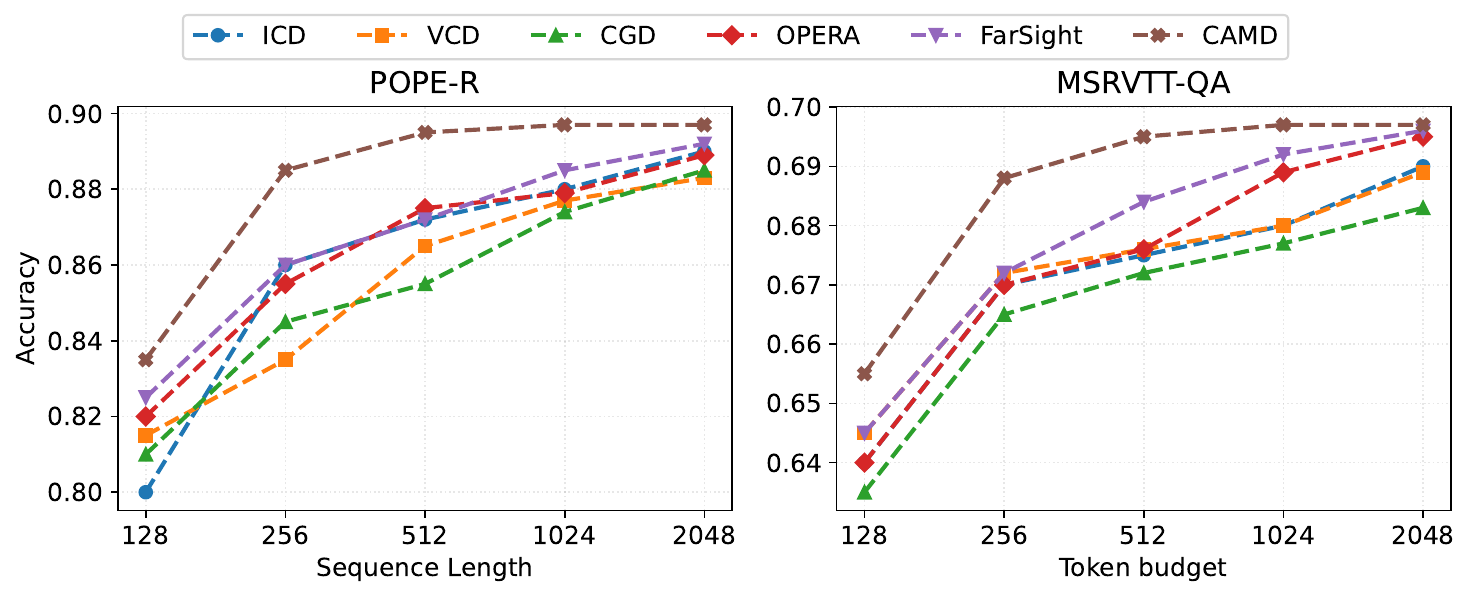}
    \vspace{-0.2in}
    \caption{Sequence Length vs. Performance with different decoding methods on POPE-R (left) and MSRVTT-QA (right).}
    \label{fig:ex_efficiency}
\end{figure}

\begin{figure}[t]
    \centering
    \includegraphics[width=\linewidth]{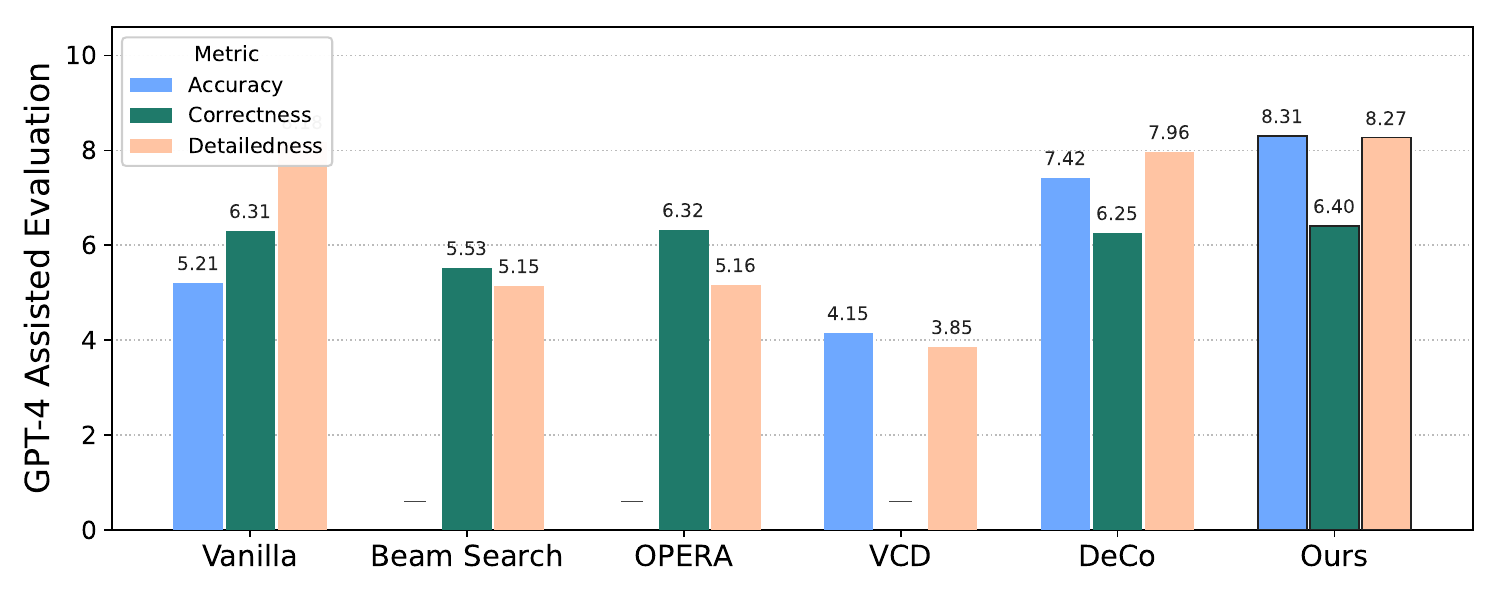}
    \vspace{-0.2in}
    \caption{Performance of different methods on GPT-4 assisted evaluation. Note that some values were not reported in the original papers; we have replaced them with ``-'' in the figure.}
    \label{fig:gpt_4o_evalution}
\end{figure}

\begin{figure}[t]
    \centering
    \includegraphics[width=\linewidth]{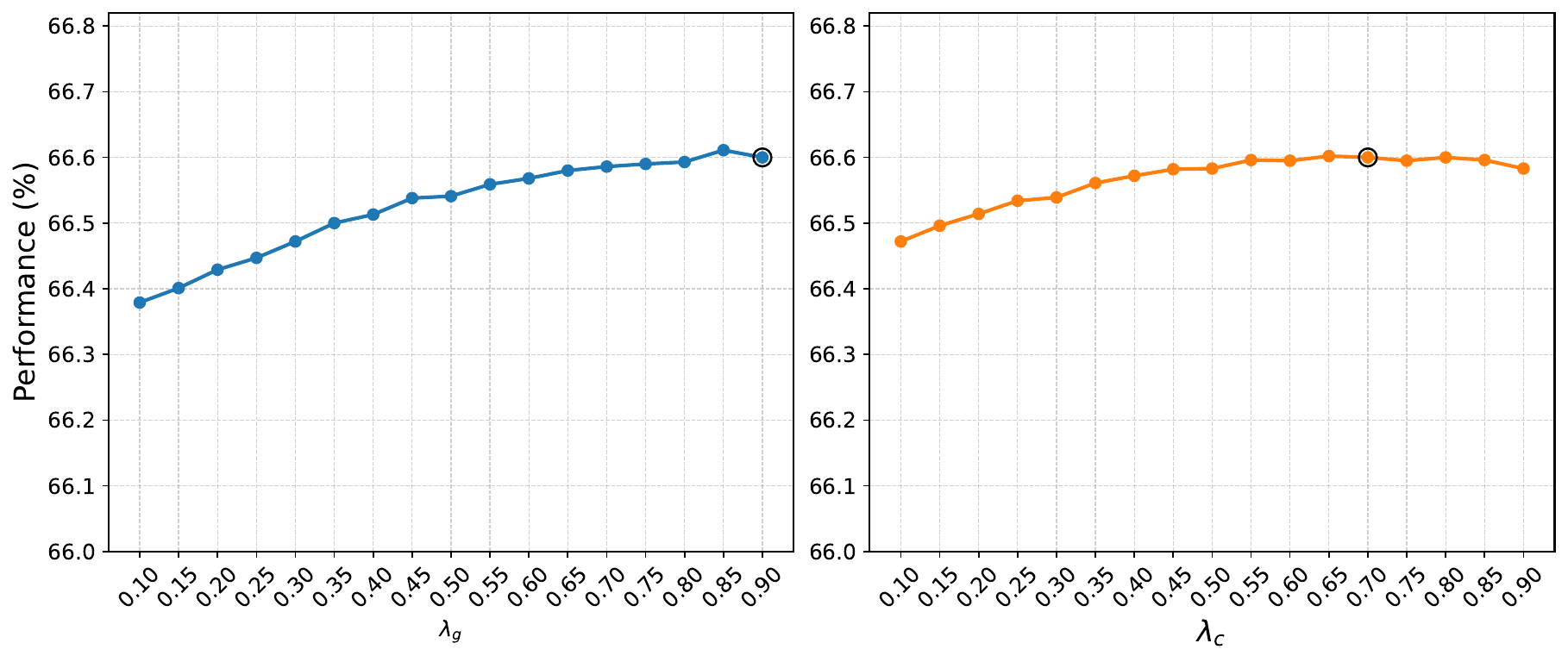}
    \vspace{-0.2in}
    \caption{Ablation study of the hyperparameters.}
    \label{fig:ablation}
\end{figure}

\section{Experiments}
\label{exp}
In this section, we conduct a series of experiments to evaluate the effectiveness and efficiency of our method. Specifically, we first introduce the experimental settings in Subsection \ref{sec:experiment_settings}. Next, we present comparative results, analyses, and ablation studies with various benchmarks in Subsection \ref{sec:experiment_comparison}. 
More details are provided in \textbf{Appendices B-E}.

\subsection{Experimental Settings}
\label{sec:experiment_settings}

\noindent\textbf{Benchmark Datasets.}
We evaluate on various benchmarks following \cite{tang2025seeing}. For images, we consider (1) comprehensive benchmarks, i.e., MMBench~\cite{liu2025mmbench}, LLaVA$^{\mathrm{W}}$~\cite{liu2024visual}, and MM-Vet~\cite{yu2023mm}; (2) general VQA benchmarks, i.e., VizWiz~\cite{gurari2018vizwiz} and SQA~\cite{lu2022learn}; and (3) hallucination-focused benchmarks, i.e., POPE~\cite{li-etal-2023-evaluating} and CHAIR~\cite{rohrbach2018object}. For video, we perform zero-shot evaluations on MSRVTT-QA~\cite{kim2023provable}, MSVD-QA~\cite{Xu2017VideoQA}, and ActivityNet-QA~\cite{Zhou2017TowardsAL}, and use the Video-Based Text Generation Benchmark for quantitative analysis~\cite{maaz2023video}. For GPT-4-assisted evaluation, we compare model-generated captions on images, with GPT-4o rating on accuracy (A), correctness (C), and detailedness (D). More details are provided in \textbf{Appendix B}.

\noindent\textbf{Baselines.}
We evaluate six representative MLLMs across image and video tasks, including InstructBLIP~\cite{dai2023instructblipgeneralpurposevisionlanguagemodels}, LLaVA-1.5~\cite{liu2024improved}, VILA~\cite{lin2023vila}, Video-LLaMA2~\cite{damonlpsg2024videollama2}, Chat-UniVi~\cite{jin2024chat}, and Video-LLaVA~\cite{lin2023video}. 
Besides, we also compare the most recently proposed decoding-based methods, including ICD~\cite{wang2024mitigating}, VCD~\cite{leng2024mitigating}, CGD~\cite{deng2024seeing} OPERA~\cite{huang2024opera}, and FarSight~\cite{tang2025seeing}.
Further details are provided in \textbf{Appendix C}.

\noindent\textbf{Implementation Details.} 
CAMD is a plug-and-play decoding wrapper that requires only the candidate outputs at decoding checkpoints without training. We set $\lambda_g=1$, $\lambda_c=0.3$; $\tau$ and $\delta$ as $0.90$ and $0.05$; clustering similarity threshold with default $0.85$. All the results are reported as the average over five runs on the A100 GPU clusters.

\subsection{Results}
\label{sec:experiment_comparison}
To evaluate the effectiveness of CAMD, we conduct experiments on several benchmarks. More details and results are provided in the \textbf{Appendix} due to space limitations.

\noindent\textbf{Consistently achieves better overall performance.}
To evaluate image and video understanding ability of CAMD, we compare models augmented with the CAMD extension against several recently proposed decoding methods. \textbf{Table \ref{tab:ex_image_result}} shows that integrating CAMD yields an average improvement of +3.5\% on the Comprehensive and General VQA tasks and produces substantial gains on hallucination metrics by more than +5\% compare with the original models; it also achieves the best results across baselines. In zero-shot video QA (\textbf{Table \ref{tab:ex_video_result}}), CAMD consistently improves performance across three major benchmarks and seven key evaluation dimensions with more than 2\%. The results demonstrate the advantages and effectiveness of CAMD.

\noindent\textbf{Consistently achieves better efficiency.} 
To evaluate the efficiency of CAMD, we compare it against various baselines on two representative benchmarks, i.e., POPE-R for the image hallucination test and MSRVTT-QA for zero-shot video QA. 
For each method, we sweep token budgets $\{128,256,512,1024,2048\}$ and measure accuracy under the same model checkpoint and prompt. 
More details are shown in the Appendix. 
From \textbf{Figure \ref{fig:ex_efficiency}}, we can observe that CAMD reaches comparable or superior peak performance at substantially lower token budgets than competing decoding methods. 
Crucially, these efficiency gains do not increase hallucination: the POPE/CHAIR scores of CAMD are comparable to or better than the baselines. These results demonstrate the effectiveness of our method.

\noindent\textbf{Performance of GPT-4o assisted evaluation.} We use GPT-4o as an automatic evaluator to measure the perceived quality of generated captions, following the protocols in \cite{huang2024opera, leng2024mitigating}.
\textbf{Figure \ref{fig:gpt_4o_evalution}}
shows that CAMD achieves the strongest results: it yields the lowest PPL and the highest GPT-4o ratings for grammar, fluency, and naturalness, producing consistently more coherent and natural outputs.

\noindent\textbf{Ablation Study.} 
We perform a hyperparameter sweep for the two weighting terms in the evidence score, i.e., $\lambda_g$ and $\lambda_c$
Concretely, we search $\lambda_g$ and $\lambda_c$ over $[0.1,0.9]$ with step size 0.05. Both alignment and coherence terms are normalized to $[0,1]$ on the validation set before weighting. As shown in \textbf{Figure~\ref{fig:ablation}}, the best validation performance is obtained at $\lambda_g=0.9$ and $\lambda_c=0.7$, also our configuration.

\noindent\textbf{Quantitative Analysis.}
We visualize the responses of LLaVA-1.5 under different methods and settings. As shown in \textbf{Appendix E}, CAMD produces more accurate, verifiable answers. Combined with the aforementioned quantitative analysis, while matching or exceeding baseline accuracy CAMD requires fewer sampling rounds and a smaller token budget, and its hallucination metrics are no worse than the baselines. Therefore, CAMD reduces inference cost while increasing answer precision and reliability.

%% file: sec/6_conclusion.tex
\section{Conclusion}
\label{sec:conclusion}

In this paper, we provide a theoretical framework connecting the sampling coverage, instance difficulty, and residual risk of MLLMs. It shows that heavy-tailed difficulty makes fixed-budget decoding inefficient for hard multimodal instances. Building on this insight, we propose \emph{Coverage-Aware Multimodal Decoding} (CAMD), which allocates computation adaptively through evidence-weighted scoring, posterior coverage estimation, and Bayesian adaptive sampling to achieve more reliable inference. A Dirichlet-based multi-cluster posterior enables modeling semantic hypotheses and reweighting tokens without manual thresholds. Experimental results conducted on various benchmark datasets and tasks demonstrate that CAMD consistently improves the performance and efficiency of MLLMs.